\documentclass[conference]{IEEEtran}
\usepackage{framed}
\usepackage[justification=centering]{caption}
\usepackage[dvipsnames]{xcolor}
\usepackage{listings}
\usepackage{balance}
\usepackage{caption}
\usepackage{colortbl}
\usepackage[final=true, debug=true, pdftex]{hyperref}
\usepackage{breakurl}
\pdfoutput=1

\usepackage{etoolbox}

\makeatletter

\pretocmd{\NAT@citex}{%
  \let\NAT@hyper@\NAT@hyper@citex
  \def\NAT@postnote{#2}%
  \setcounter{NAT@total@cites}{0}%
  \setcounter{NAT@count@cites}{0}%
  \forcsvlist{\stepcounter{NAT@total@cites}\@gobble}{#3}}{}{}
\newcounter{NAT@total@cites}
\newcounter{NAT@count@cites}
\def\NAT@postnote{}

\def\NAT@hyper@citex#1{%
  \stepcounter{NAT@count@cites}%
  \hyper@natlinkstart{\@citeb\@extra@b@citeb}#1%
  \ifnumequal{\value{NAT@count@cites}}{\value{NAT@total@cites}}
    {\ifNAT@swa\else\if*\NAT@postnote*\else%
     \NAT@cmt\NAT@postnote\global\def\NAT@postnote{}\fi\fi}{}%
  \ifNAT@swa\else\if\relax\NAT@date\relax
  \else\NAT@@close\global\let\NAT@nm\@empty\fi\fi
  \hyper@natlinkend}
\renewcommand\hyper@natlinkbreak[2]{#1}

\patchcmd{\NAT@citex}
  {\ifNAT@swa\else\if*#2*\else\NAT@cmt#2\fi
   \if\relax\NAT@date\relax\else\NAT@@close\fi\fi}{}{}{}
\patchcmd{\NAT@citex}
  {\if\relax\NAT@date\relax\NAT@def@citea\else\NAT@def@citea@close\fi}
  {\if\relax\NAT@date\relax\NAT@def@citea\else\NAT@def@citea@space\fi}{}{}

\makeatother
\makeatletter
\def\lst@numbersymbol{}
\lst@Key{numbersymbol}{}{\def\lst@numbersymbol{#1}}
\lst@Key{numbers}{none}{%
    \let\lst@PlaceNumber\@empty
    \lstKV@SwitchCases{#1}%
    {none&\\%
     left&\def\lst@PlaceNumber{\llap{\normalfont
                \lst@numberstyle{\thelstnumber\lst@numbersymbol}\kern\lst@numbersep}}\\%
     right&\def\lst@PlaceNumber{\rlap{\normalfont
                \kern\linewidth \kern\lst@numbersep
                \lst@numberstyle{\lst@numbersymbol\thelstnumber}}}%
    }{\PackageError{Listings}{Numbers #1 unknown}\@ehc}}
\makeatother

\definecolor{darkspringgreen}{rgb}{0.09, 0.45, 0.27}
\definecolor{dartmouthgreen}{rgb}{0.05, 0.5, 0.06}

\usepackage{tikz}

\usepackage{lipsum} 
\usepackage{dingbat}
\usepackage{pifont}
\usepackage{amssymb}
\makeatletter
\def\@copyrightspace{\relax}
\makeatother

\begin{document}
\definecolor{carnelian}{rgb}{0.7, 0.11, 0.11}

\title{Mobile Robot Navigation on Partially Known Maps using a Fast A$^\ast$ Algorithm Version}


\author{\IEEEauthorblockN{Paul Muntean}
\IEEEauthorblockA{Technical University of Munich, Germany\\
paul.muntean@in.tum.de
}}

\maketitle
\begin{abstract}
Mobile robot navigation in total or partially unknown
environments is still an open problem. The path planning 
algorithms lack completeness and/or performance. Thus,
there is the need for complete (i.e., the algorithm determines in finite time either a solution or
correctly reports that there is none) and performance (i.e., with low computational complexity) oriented algorithms
which need to perform efficiently in real scenarios.

In this paper, we evaluate the efficiency of two versions of 
the A$^\ast$ algorithm for mobile robot navigation inside
indoor environments with the help of two software applications and the Pioneer 2DX robot.
We demonstrate that an improved version of the A$^\ast$ algorithm
which we call the \emph{fast} A$^\ast$ algorithm
can be successfully used for indoor mobile robot navigation.
We evaluated the A$^\ast$ algorithm first, by
implementing the algorithms in source code and by testing them on
a simulator and second, by comparing two operation modes of the \emph{fast} A$^\ast$ algorithm
w.r.t. path planning efficiency (i.e., completness) and performance (i.e., time need to complete the path traversing)
for indoor navigation with the Pioneer 2DX robot.
The results obtained with the \emph{fast} A$^\ast$ algorithm are
promising and we think that this results can be 
further improved by tweaking the algorithm and by 
using an advanced sensor fusion approach 
(i.e., combine the inputs of multiple robot sensors) for 
better dealing with partially known environments.

\end{abstract}

\IEEEpeerreviewmaketitle

\section{Introduction}
\label{Intro}

Motion planning---also known as the navigation problem or the piano mover's problem---is a term used in robotics for the process of breaking down a desired movement task into discrete motions that satisfy movement constraints and possibly optimize some aspect of the movement.
Motion planning has several robotics applications, such as: 
\textit{(i)} robot navigation,
\textit{(ii)} automation,
\textit{(iii)} the driver-less car,
\textit{(iv)} robotic surgery,
\textit{(v)} digital character animation,
\textit{(vi)} protein folding,
\textit{(vii)} safety and accessibility in computer-aided architectural design,
\textit{(viii)} UCAV Path Planning~\cite{ucav:z}, etc.

A basic motion planning problem is to produce a continuous motion that connects 
a start configuration S and a goal configuration G, while avoiding 
collision with known obstacles. The robot and obstacle geometry
is described in a 2D or 3D workspace, while the motion is 
represented as a path in (possibly higher-dimensional) 
configuration space (describes the pose of the robot, 
and the configuration space C is the set of all possible configurations).

\emph{\textbf{Problem statement:}} Recent studies show that every day activity of people in
cities and countries living in the modern society is rapidly 
increasing~\cite{Siegwart:r} in such a way that efficient navigation of people movement
is needed. Researchers have
tried to come with new and better navigation approaches in
the past as for example Jones~\cite{jonnes:j}.
These approaches lack efficiency or applicability to mobile 
robot navigation in real path planning environments.

\emph{\textbf{Available solutions:}} Path planing w.r.t.
low-dimensional problems can be addressed using: 
\textit{(i)} grid-based approaches which overlay a grid on a configuration space and assume that each configuration is identified by a grid point. 
At each grid point, the robot is allowed to move to adjacent grid points as long as 
the line between them is completely contained within $C_{free}$ (the set of configurations that avoids collision with obstacles is called the free space $C_{free}$) 
(this is tested with collision detection),
\textit{(ii)} interval-based search which is similar to grid-based search approaches 
except that they generate a paving covering entirely the configuration space 
instead of a grid~\cite{jaulin:j},
\textit{(iii)} geometric algorithms which are used to point robots among polygonal 
obstacles based on a visibility graph, cell decomposition
and translating objects among obstacles
using the Minkowski sum~\cite{minkowski:addition},
\textit{(iv)} potential fields which are used to treat the robot's configuration as a point in a potential field that combines attraction to 
the goal and repulsion from obstacles. 
The resulting trajectory represents the new path which is computed fast. However, they can become trapped in local minima of the potential field, and fail to find a path,
\textit{(v)} sampling-based algorithms which represent the configuration space with a road-map of sampled configurations. A basic algorithm samples N configurations in C, and retains those in $C_{free}$ to use as milestones. A road-map is then constructed that connects two milestones P and Q if the line segment PQ is completely in $C_{free}$.
Most notable algorithms are the A* and D* algorithms which can rapidly explore random trees and probabilistic road-maps.

A motion planning algorithm is said to be complete if the 
planner determines in finite time either a solution or 
correctly reports that there is none. Most complete 
algorithms are geometry-based. 
Resolution completeness is the property that the planner 
is guaranteed to find a path if the resolution of an 
underlying grid is fine enough. Most resolution complete 
planners are grid-based or interval-based.
Probabilistic completeness states that, as more ``work” is performed, 
the probability that the planner fails to find a path (if one exists)
asymptotically approaches zero. 
The performance of a probabilistically 
complete planner is measured by the rate of convergence.
Incomplete planners do not always produce a feasible 
path when one exists. 
The performance of a complete planner is assessed by 
its computational 
complexity computed using the big $\mathcal{O}$ notation.

\emph{\textbf{Deficiencies of available solutions:}} In summary, existing path planning algorithms lack in
determining a path when one exists or they need to much time
to compute one. Thus, the main limitations of these 
algorithms are related to completeness and/or performance.
Thus, in this work wee seek for a suited robot path planning algorithm
which is complete and performant.

\emph{\textbf{Our idea:}} Our insight is that an improved A$^\ast$ algorithm (we call this the \emph{fast} A$^\ast$ algorithm) 
can be efficiently used for path planning of real robots in a partially known environment.
We evaluated two versions of the A$^\ast$ algorithm and
presented the results obtained with the Pioneer 2DX robot~\cite{pioneer2dx:robot}.
The communication (closed loop) between our PC and the real robot was 
achieved by sending real-time navigation commands via a wireless connection
based on the Lantronix WiBox~\cite{lantronix}.
Note that during the experiments the Pioneer 2DX robot used only the 
ultrasonic sensors in order to partially reconstruct a map of the partially
known (containing unknown obstacles) environment---not mapped on the initial on-line mode navigation map.

In this paper, we address the problem of efficient and complete motion planning of
a three wheeled mobile robot by implementing two algorithms
(the A$^\ast$ algorithm and the \emph{fast} A$^\ast$ algorithm) and comparing the 
efficiency (with focus on completness and performance) of this two approaches
on a path planning algorithm simulator and afterwards with the real Pioneer 2DX robot.

\emph{\textbf{Our contributions:}} In summary, the main contributions are:
\begin{itemize}
 \item We develop an improved version of the A$^\ast$ algorithm which proves to
       be faster in offline testing (with a software simulator) and efficient in real environments 
       when tested with the real Pioneer 2DX robot.
       
 \item We implement two applications: first, a simulator used for path planning 
       simulation in offline mode (not with the real robot) and assessed the performance and completness of the
       A$^\ast$ algorithm and of the \emph{fast} A$^\ast$ algorithm and second, a path planning application
       used in online-mode (with the Pioneer 2DX mobile) to navigate
       him through a partially known map using only the \emph{fast} A$^\ast$ algorithm in two different operation modes.
       
 \item We demonstrate that the \emph{fast} A$^\ast$ algorithm
       is effective when tested with the Pioneer 2DX mobile robot inside a partially known 
       indoor environment\footnote{ \textbf{Demo movie available:}~\url{https://goo.gl/OYXMDy}}.

\end{itemize}

The remainder of this paper is organized as follows.
Section,~\ref{Related} highlights background work.
Section,~\ref{Algorithm} presents the A$^\ast$ algorithm.
Section,~\ref{Implementation} highlights implementation details.
Section,~\ref{Experiments} depicts experiments results.
Finally, in Section~\ref{Conclusion and Future Work} we conclude and present future work.

\section{Background}
\label{Related}
\subsection{Brief Routing History}
In the 1970 scientists started research on routing algorithms for moving chess pieces on a chess-board and on how to efficiently move fragments on a puzzle map~\cite{eklund:p}.
As a consequence the research on routing algorithms started.
The main reason for starting the research in the area of routing algorithms
was that these problems can be easily abstracted and further on
the results can be applied to more complex fields of study such as robot navigation.
Thus, with the development of path finding, several new classical routing algorithms 
have emerged at that time with the goal to generate better routing results. 

The Dijkstra algorithm is the most famous algorithm. The algorithm evaluates the moving cost from one node to any other node and sets the shortest moving cost as the connecting cost of two nodes~\cite{eklund:p}.
Around the same period the Best First Search (BFS) algorithm
was introduced. 
The BFS is different from the Dijkstra algorithm, since the BFS estimates the distance from the current position to goal position and it chooses the next step that is more closer to the goal position~\cite{amit:game}. 

As the complexity of the path finding scenarios was growing the path finding algorithms had to be improved in order to meet new
requirements as for example 3D maps.

\subsection{The A* Algorithm and Extensions}

As response to the new path planning requirements the A$^\ast$ algorithm appeared. The goal of the new A$^\ast$ algorithm
is path planning efficiency.
The A$^\ast$ algorithm is a BFS algorithm which uses huge amounts of memory in order to keep track of the data related to the current proceeding nodes~\cite{Nelson:p}. 
The A$^\ast$ algorithm tries to combine the advantages offered by the Dijkstra algorithm and the BFS algorithm. The A$^\ast$ algorithm tries during each new movement to take the shortest step and tries to determine if the step lies on the direction from source to target~\cite{jonnes:j}.
The disadvantage of the A$^\ast$ algorithm is that it uses large amounts of memory in order to store the path planning environment.

The A$^\ast$ algorithm proved to have its limitations and in response new methods of using the A$^\ast$ algorithm appeared. The bidirectional A$^\ast$ algorithm~\cite{Nelson:p} is used in order to reduce the time cost of the A$^\ast$ algorithm. 
The most important difference of the
bidirectional A$^\ast$ algorithm w.r.t. the classical A$^\ast$ algorithm
(which is searching from the source to the target location) is that
it can search from source to target and vice-versa. The path search stops immediately when the two directional searching processes meet each other.

The Iterative Deepening A$^\ast$ (IDA$^\ast$)~\cite{korf} is a space-efficient version of the A$^\ast$ algorithm, which
suffers from cycles in the search space 
(it uses no storage), repeated visits to states 
(the overhead of iterative deepening), 
and a simplistic traversal of the search tree.
Since it is a depth-first search algorithm, its 
memory usage is lower than in A$^\ast$, but unlike ordinary 
iterative deepening search, it concentrates on exploring the most promising nodes and thus does not 
go to the same depth everywhere in the search tree. Unlike A$^\ast$, IDA$^\ast$ 
does not utilize dynamic programming and therefore often ends up 
exploring the same nodes many times~\cite{ida:wiki}.

Routing in three dimensions (3D) is much more complex than
under two space dimensions, thus the traditional A$^\ast$ algorithm should be improved in order 
to meet the additional routing requirements. 
The three dimensional A$^\ast$ algorithm
has emerged as a response for better dealing with 3D environments.
The three dimensional A$^\ast$ algorithm was obtained by adding several modifications to the A$^\ast$ algorithm in order to be used for computing navigation paths in 3D maps (e.g., the path planning of a cart in a mine system which has multiple levels).

Furthermore, a frequently used approach for solving simple three dimensional path planning problems is to map the three dimensional map
into a two dimensional expression.
In this way the traditional
A$^\ast$ algorithm can be used for solving the path planning~\cite{Makanae:k} in 3D environments.
Note that this technique of mapping 3D maps to 2D maps is working for path planning in simple 3D scenarios---reduced set of constrains. In complex scenarios this mapping method can not be used and thus more complex approaches are needed.

\section{The A$^\ast$ algorithm} 
\label{Algorithm}
In this section, we briefly describe the main parts of the A$^\ast$ algorithm.
The A$^\ast$ algorithm~\cite{astar:j} uses the BFS algorithm
in order to find the least-cost path from a given initial node
to one goal node (the last position could be a single or multiple nodes). 
It uses a distance-plus-cost heuristic function 
(usually denoted by \emph{f(x)}) to determine the order in
which the search visits nodes in the node tree. 
The distance-plus-cost
heuristic \emph{f(x)}) is expressed as sum of two functions:
(a) the path-cost function, represents the cost from the starting node to the current node
(usually denoted \emph{g(x)}) and 
(b) an admissible ``heuristic estimate" used
to model an estimated from the current position/node to the 
the goal position/node (usually denoted with \emph{h(x)}).
The distance-plus-cost heuristic function can be framed as follows.

\begin{figure}[ht!]
    \centering
    \noindent \hfill $f(x) = g(x) + h(x)$    \hfill (1)                           
    \label{fig:awesome_image}
\end{figure}

An important constraint is that the \emph{h(x)} component of \emph{f(x)}
must be an admissible heuristic---briefly this means that it is important to not
overestimate the distance from current node to the goal node.
The \emph{g(x)}
component of \emph{f(x)} represents the
total cost from the start node and not only
the cost from the previously expanded (visited) node.
In case of determining the shortest distance between two
locations (nodes) it is known that the straight line is the shortest
distance. In case of routing, \emph{h(x)} could be represented as a straight-line
distance from current position to the goal position.
Next we impose the following constraint on \emph{h(x)}.

\begin{figure}[ht!]
    \centering
    \noindent \hfill  $h(x) \le d(x,y)+h(y)$     \hfill (2)   
    \label{fig:awesome_image}
\end{figure}

The mathematical expression (2) imposes that every edge represented by \emph{x}
and \emph{y} belonging to a graph where \emph{d(x,y)} represents
the length of the given edge results in an \emph{h{x}} which is consistent or monotone.
Furthermore, (2)
guarantees that one node is processed only once and in this
case the implementation of the A$^\ast$ is more efficient. 
In this case running the A$^\ast$ algorithm is similar to 
running the Dijkstra's algorithm having the cost reduced.
Next we impose the following constraint on the length of a graph edge.

\begin{figure}[ht!]
    \centering
        \noindent \hfill   $d'(x,y) = d(x,y) - h(x) + h(y)$        \hfill (3)   
    \label{fig:awesome_image}
\end{figure}

The A$^\ast$ algorithm is an \textit{informed}
search algorithm. A particularity of \textit{informed} search algorithms is to
search for the routes (paths) that appear to be most likely to lead to the
goal position. Note that the A$^\ast$ algorithm differs from the greedy
best-first search algorithm because it takes into consideration the already 
travelled distance.
The process of finding the path from a starting position to a
target position by using the A$^\ast$ algorithm is repetitive
and ends when the current visited node is equal to the target
node or when the target position is reached. 
During graph nodes traversing the A$^\ast$ algorithm
follows a path from the lowest known path based on
keeping a priority queue of all alternate path segments along
the path. 
When an edge of the path is
traversed which has a higher cost than another previously encountered path 
segment then it immediately abandons the current path segment (having
higher cost) and continues with the lower-cost path segment.

Note that each node points
to his parent node and in case of encountering a solution the path
can be easily returned and added to a list of optimal paths.
The A$^\ast$ algorithm can be implemented based on a loop
in which a repeated check of a node (e.g., \emph{n}) is performed
having the lowest, \emph{f(n)} value from an \textit{open} list of nodes.
The analyzed node \emph{n} is considered to be the most likely candidate to 
be part of the optimal path.
If \emph{n} is the target node then only one 
backtracking step has to be performed in order to return the 
obtained solution.
If \emph{n} is not the final node then \emph{n} has to be
removed from the previously mentioned \textit{open}
list and introduced into another list which we call the \textit{closed} list.
The next step consists of generating all possible successor nodes of \emph{n}.

However, in order to efficiently implement the A* algorithm
it is important to take into consideration that, \emph{g(n)} represents the
cost to get the distance from the initial node to the \emph{n}'th node;
\emph{h(n)} is an estimate and represents the cost of getting 
the distance from node \emph{n} to a goal node.
The equation \emph{f(n) = g(n) + h(n)} represents an
estimate of the best solution that contains the node \emph{n}.

\section{The fast A$^\ast$ Algorithm Implementation}
\label{Implementation}
In this section, we first, present implementation details of our \emph{fast} A$^\ast$ algorithm
and
second, we present implementation details of our two tools.

\subsection{The \emph{Fast} A$^\ast$ algorithm Implementation}
\begin{lstlisting}[caption=The A$^\ast$ algorithm---informal description, 
language=C, morekeywords={open, closed, goal, start, if, return, for, then, continue},
label=algorithmpseudo]
initialize the open list
initialize the closed list
initialize goal node  <@\textcolor{OliveGreen}{\textbf{// this is the target node}}@>
initialize start node <@\textcolor{OliveGreen}{\textbf{// add the node to the open}}@>
while open list is not empty {
 get node n from the open list with the lowest f(n)
 add n to the closed list
  if n is equals the goal node then stop;
     return solution;
   generate each successor node n' of n;
   for each successor node n' of n {
      set the parent of n' to n;
      <@\textcolor{OliveGreen}{\textbf{// heuristic estimate distance to goal node}}@>
      set h(n') 
      set g(n') = g(n) + cost from n to get to n' 
      set f(n') = g(n') + h(n')
     if n' contained in open and the existing node 
          is as good or better then discard n' 
          and continue;
     if n' is contained in closed and the existing 
          node is as good or better then discard
          n' and continue;
     remove all occurrences of n' from open and
     closed and add n' to the open list;
   }
}
<@\textcolor{OliveGreen}{ \textbf{// if we searched all reachable nodes}}@>
<@\textcolor{OliveGreen}{ \textbf{// and still have not found a solution then return}}@>
return failure; 
\end{lstlisting}

The algorithm depicted in Listing~\ref{algorithmpseudo} has as input the \emph{open}
list containing all nodes which can be visited. The \emph{open} list
is implemented as a balanced binary tree sorted based on \emph{f} values,
with tie-breaking in favor of higher \emph{g} values.
The tie-breaking mechanism results in the goal state being
found on average earlier in the last~\emph{f}~value pass.
In addition to the standard \emph{open} and \emph{closed} lists,
marker arrays are used for finding in constant time whether a 
state (node) is in the \emph{open} or
\emph{closed} list. 
We use a ``lazy-clearing” scheme in order to avoid having
to clear the marker arrays at the beginning of each search. 
Each path search is assigned a unique increasing \emph{ID}
that is then used to label array entries relevant for the
current performed search.
Note that the \emph{closed} list can be omitted (yielding a tree search algorithm)
if a solution is guaranteed to exist or if the algorithm is
adapted such that new nodes are added to the open list only if
they have a lower \emph{f} value at any previous iterations.
The \emph{fast} A$^\ast$ algorithm keeps an open 
node in a priority queue such that it avoids closing this 
list which normally happens when the node is 
removed. Thus the search process is speeded up.
Additionally, we tested different algorithm heuristics,
metrics, the allowance of diagonals traversing of map tiles and came
up with an efficient set of settings for our algorithm.
These characteristics represent the main differences 
of our \emph{fast} A$^\ast$ w.r.t. the A$^\ast$ algorithm.
Thus, our \emph{fast} A$^\ast$ algorithm implementation
provides an order of magnitude performance improvement over the
standard textbook A$^\ast$ implementation~\cite{Siegwart:r}.
Note that similar versions of algorithms to ours are successfully 
used for as path planning algorithms inside video games 
(e.g., Counter-Strike video game~\cite{counterstrike:videogame}). 

\subsection{Offline and Online Tools}

The \emph{fast} A$^\ast$ algorithm depicted in Listing~\ref{algorithmpseudo} was implemented and tested with
two applications (\textit{(i)} offline mode and \textit{(ii) online mode}).
First, \textit{(i)} a C\# based application was developed (see the GUI in 
Figure~\ref{Map used for testing}) used for simulating the 
A$^\ast$ algorithm and the \emph{fast} A$^\ast$ algorithm in offline
mode with different parameter configurations.
Second, \textit{(ii)} a path planning application (used to remotely steer/control via
the Lantronix WiBox the Pioneer 2DX robot along a navigation path) was 
developed based on the Java based Saphira API~\cite{saphira:r} 
using the Java Native Interface (JNI) and the 
Aria API~\cite{aria:j}.
Finally, in order to determine the time needed for calculating an optimum
path from the starting position to the target position the offline application
used a high resolution timer which was implemented
using the Windows OS \emph{kernel32.dll} library.  

\section{Experiments}
\label{Experiments}
In this section, we evaluate (i.e., offline in Section~\ref{The a vs a fast} and 
online in Section~\ref{Path Planning with the Pioneer robot}) the A$^\ast$ algorithm 
and the \emph{fast} A$^\ast$ algorithm
in order to determine which would better fit to be used in online-mode
with the Pioneer 2DX robot.

\subsection{The A$^\ast$ Algorithm vs. Fast A$^\ast$ Algorithm in Offline Mode}

\label{The a vs a fast}
\begin{figure}[ht!]
    \centering
    \includegraphics[width=0.485\textwidth,natwidth=531,natheight=425]{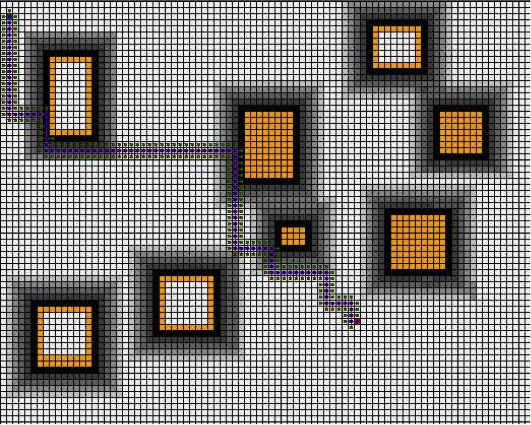}
    \caption{The 2D map used to test the A$^\ast$  and the \emph{fast} A$^\ast$ algorithms. 
    Start position was always the upper left corner selected while
    final destination was selected around the center of the map.}
    \label{Map used for testing}
\end{figure}

Figure~\ref{Map used for testing} represents the map used for calculating the 
times for each of the run-times during offline simulation of the A$^\ast$ algorithm and
\emph{fast} A$^\ast$ algorithm. The rectangles filled or having orange border depicted in 
Figure~\ref{Map used for testing} represent obstacles (not passable map areas). 
The path depicted in Figure~\ref{Map used for testing} with interconnected blue tiles
from the top left corner towards the middle of the map represents a valid robot navigation path.
The valid path avoids obstacles depicted in Figure~\ref{Map used for testing}
with rectangles having an orange border and additionally several 
borders depicted with different levels of grey color.
Note that as darker the grey color tone is (in the map tiles) as forbidden the area
is for the path planning algorithm. 
Thus, the algorithm tries to avoid these areas as much as possible.

Note that we conducted each run for a set of parameters by increasing the heuristic number 
(\textbf{Heuristic \textbf{\#}} see Table~\ref{Test results for the A algorithm}
and Table~\ref{Test results for the A fast algorithm}) from \emph{0}
to \emph{n} as long as the run-time calculated in seconds was decreasing.
The first time we noticed that the run-time was increasing we stopped the test run and we selected another
formula and repeated the experiment by starting with the heuristic number \emph{0}.
The experiments were conducted in this manner in order to find out which is the best configuration
for the set of parameters used inside the two path planning algorithms. 
Note that in a real scenario (the environment can constantly change) path planning computations 
need to be performed with a higher rate (e.g., in our opinion less than 100 milliseconds).

\begin{table}[h!]
\centering
\caption{Test results of the A$^\ast$ algorithm}
\label{Test results for the A algorithm}
\begin{tabular}{|c|c|c|c|c|} \hline
\textbf{\#}    &\textbf{Heuristic \textbf{\#}}     & \textbf{Diagonals}                         &\textbf{Formula} & \textbf{Time [sec]}  \\ \hline 
1              &0                      &\checkmark                                  &\emph{m}         &$\geq$30 \\   
2              &1                      &\checkmark                                  &\emph{m}         &1.34     \\   
3              &2                      &\checkmark                                  &\emph{m}         &0.01     \\  
4              &3                      &\checkmark                                  &\emph{m}         &0.03     \\  \hline 
5              &0                      &\checkmark                                  &\emph{M(x,y)}    &10.68    \\ 
6              &1                      &\checkmark                                  &\emph{M(x,y)}    &2.74     \\  
7              &2                      &\checkmark                                  &\emph{M(x,y)}    &$\geq$30 \\  \hline 
8              &0                      &\checkmark                                  &\emph{D.S.}      &$\geq$30 \\ 
9              &1                      &\checkmark                                  &\emph{D.S.}      &1.31     \\  
10             &2                      &\checkmark                                  &\emph{D.S.}      &0.03     \\  
11             &3                      &\checkmark                                  &\emph{D.S.}      &0.34     \\  \hline 
12             &0                      &\checkmark                                  &\emph{E}         &25.30    \\  
13             &1                      &\checkmark                                  &\emph{E}         &2.20     \\  
14             &2                      &\checkmark                                  &\emph{E}         &$\geq$30 \\  \hline 
15             &0                      &\checkmark                                  &\emph{SQR}       &25.96    \\  
16             &1                      &\checkmark                                  &\emph{SQR}       &$\geq$30 \\  \hline 
\textbf{Total} &-                      &-                                           &-                &219.94   \\  \hline 
\end{tabular} 
\end{table} 

\begin{table}[h!]
\centering
\caption{Test results of the \emph{fast} A$^\ast$ algorithm}
\label{Test results for the A fast algorithm}
\begin{tabular}{|c|c|c|c|c|} \hline
\textbf{\#}   &\textbf{Heuristic \textbf{\#}}     & \textbf{Diagonals}     &\textbf{Formula} & \textbf{Time [sec]}  \\ \hline 
1             &0                      &\checkmark              &\emph{m}         &0.09   \\   
2             &1                      &\checkmark              &\emph{m}         &0.03   \\ 
3             &2                      &\checkmark              &\emph{m}         &0.01   \\  
4             &3                      &\checkmark              &\emph{m}         &0.003  \\  
5             &4                      &\checkmark              &\emph{m}         &0.003  \\  \hline 
6             &0                      &\checkmark              &\emph{M(x,y)}    &0.10   \\  
7             &1                      &\checkmark              &\emph{M(x,y)}    &0.04   \\  
8             &2                      &\checkmark              &\emph{M(x,y)}    &0.02   \\   
9             &3                      &\checkmark              &\emph{M(x,y)}    &0.01   \\   
10            &4                      &\checkmark              &\emph{M(x,y)}    &0.01   \\   
11            &5                      &\checkmark              &\emph{M(x,y)}    &0.01   \\  
12            &6                      &\checkmark              &\emph{M(x,y)}    &0.008  \\   
13            &7                      &\checkmark              &\emph{M(x,y)}    &0.007  \\  
14            &8                      &\checkmark              &\emph{M(x,y)}    &0.007  \\  \hline 
15            &0                      &\checkmark              &\emph{D.S.}      &0.10   \\  
16            &1                      &\checkmark              &\emph{D.S.}      &0.02   \\  
17            &2                      &\checkmark              &\emph{D.S.}      &0.01   \\   
18            &3                      &\checkmark              &\emph{D.S.}      &0.0034 \\   
19            &4                      &\checkmark              &\emph{D.S.}      &0.0037 \\  \hline 
20            &0                      &\checkmark              &\emph{E}         &0.11   \\ 
21            &1                      &\checkmark              &\emph{E}         &0.04   \\  \hline 
22            &2                      &\checkmark              &\emph{E}         &0.02   \\  
23            &3                      &\checkmark              &\emph{E}         &0.02   \\   
24            &4                      &\checkmark              &\emph{E}         &0.01   \\  
25            &5                      &\checkmark              &\emph{E}         &0.01   \\  \hline 
26            &0                      &\checkmark              &\emph{SQR}       &0.12   \\  
27            &1                      &\checkmark              &\emph{SQR}       &0.01   \\  
28            &2                      &\checkmark              &\emph{SQR}       &0.0012 \\  
29            &3                      &\checkmark              &\emph{SQR}       &0.0011 \\  
30            &4                      &\checkmark              &\emph{SQR}       &0.0015 \\  \hline 
\textbf{Total}&-                      &-                       &-                &0.83   \\  \hline 
\end{tabular}
\end{table} 

Table~\ref{Test results for the A algorithm}
and Table~\ref{Test results for the A fast algorithm}
depict with: (\textbf{\textbf{\#}}) the number of the run,
(\textbf{Heuristic \textbf{\#}}) the heuristic number which can vary between \emph{1} and \emph{n},
(\textbf{Diagonals}) if diagonals on the path were allowed (\checkmark) or not,
(\textbf{Formula}) different formulas used for the distance metric 
(e.g., \emph{m}-manhattan, \emph{M(x,y)}-Max(Dx,Dy), \emph{D.S.}-diagonal shortcut,
\emph{E}-Euclidean and \emph{SQR}-Euclidean without square) and
(\textbf{Time [sec]}) in seconds for each run.
Table ~\ref{Test results for the A algorithm}
and Table~\ref{Test results for the A fast algorithm} depict the test results
obtained with our offline simulator application designed
for testing the A$^\ast$ algorithm and the \emph{fast} A$^\ast$ algorithm separately with 
different maps. We used several algorithm configurations and tested on the 2D map
depicted in Figure~\ref{Map used for testing}.

Table~\ref{Test results for the A algorithm} 
and Table~\ref{Test results for the A fast algorithm} depict the obtained results
for the two used algorithms (A$^\ast$ and \emph{fast} A$^\ast$). Thus we observe that the \emph{fast} A$^\ast$ algorithm 
is two orders of magnitude (263 = 219 [s] / 0.83 [s], see Table~\ref{Test results for the A algorithm} 
and Table~\ref{Test results for the A fast algorithm}) faster than the A$^\ast$ 
algorithm with respect to \textbf{Total} time. 
The \emph{fast} A$^\ast$ algorithm allows to increase the heuristic number
for 8 times instead the classic A$^\ast$ allows this number
to be increased up to a maximum of three times.
The shortest time is also obtained for the \emph{fast} A$^\ast$ algorithm
which also shows that the performances of the classical A$^\ast$
algorithm can be further increased in order to gain more speed.
This is also due to algorithm implementation particularities of the \emph{fast}
A$^\ast$ algorithm which leaves the open node in the priority queue. 
In summary the obtained results show that \emph{fast} A$^\ast$ algorithm is the best fit 
for usage when performing path planning with the real Pioneer 2DX robot.

\subsection{Path Planning with the Fast A$^\ast$ Algorithm in On-line Mode}
\label{Path Planning with the Pioneer robot}
The goal of this experiment is to measure the run-times obtained for
different runs and to find out how the robot manages to follow a given path by
avoids previously unknown path obstacles. 
In this experiment we used the WiFi based application which planned and steered
the Pioneer 2DX robot in a partially known environment by using two running modes.


\begin{figure}[ht!]
    \centering
    \includegraphics[width=0.45\textwidth,natwidth=477,natheight=499]{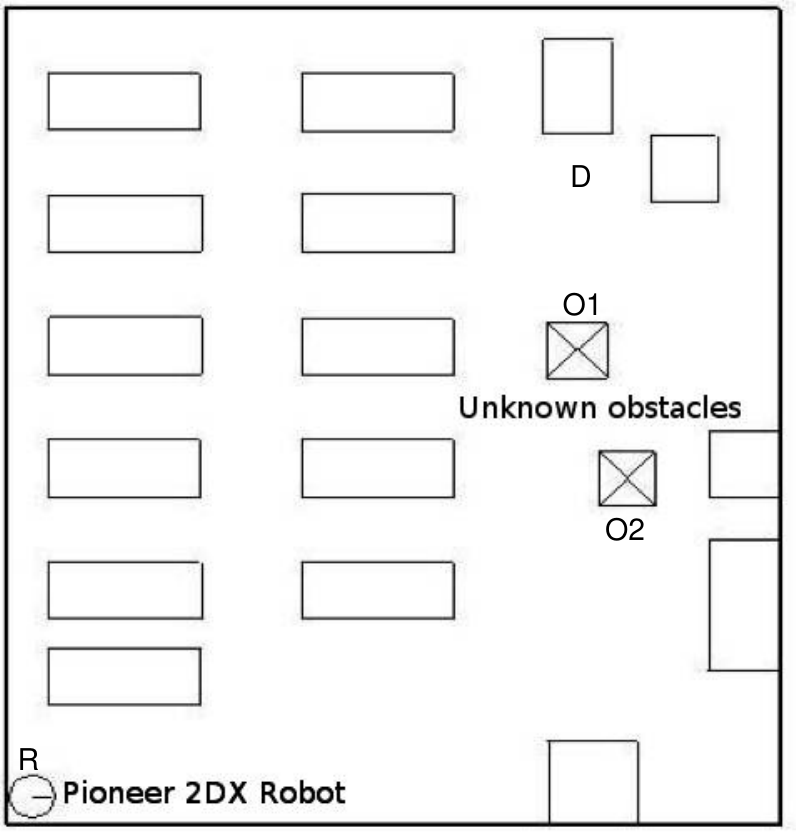}
    \caption{Second environment map used with the Pioneer 2DX. Initial robot position (R); Final robot destination (D).}
    \label{Second environment/map used to test Pioneer 2DX}
\end{figure}

\begin{figure}[ht!]
    \centering
    \includegraphics[width=0.40\textwidth,natwidth=470,natheight=508]{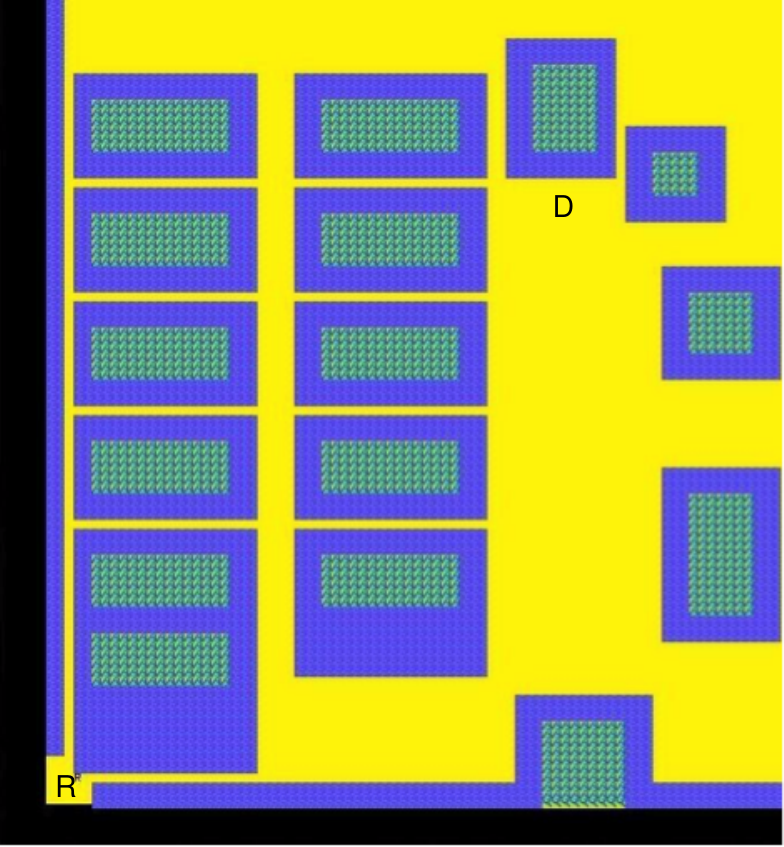}
    \caption{Environment map available in the path planner. Initial robot position (R); Final robot destination (D).}
    \label{Path planer map available in the Java back-end}
\end{figure}

Figure~\ref{Second environment/map used to test Pioneer 2DX}
depicts a partially known environment. 
Note that the rectangles with diagonals lines inside (O1 and O2) depicted in 
Figure~\ref{Second environment/map used to test Pioneer 2DX} 
represent unknown obstacles which were previously not modeled inside the 
path planning application map (Java back-end application)
depicted in Figure~\ref{Path planer map available in the Java back-end}. 
The \emph{fast} A$^\ast$ algorithm was tested on this two maps (with only and O1 and then with both O1 and O2)
with the real Pioneer 2DX robot simulator~\cite{saphira:r}
with the goal to find out if the robot can deal with partially known environments.
The experiment was performed in a room having six by eight meters and
by re-modeling it in the steering application depicted in Figure~\ref{Path planer map available in the Java back-end}.
We used for the online experiments the 14 configuration from Table~\ref{Test results for the A fast algorithm}
(i.e., \textbf{Heuristic \#} 8, \textbf{Diagonals} on (\checkmark) and \textbf{Formula} \emph{M(x,y)}).
We decided to use this configuration because it was the longest run from our experiments where the 
\textbf{Heuristic \#} number could be increased (8 times) until the search time (\textbf{Time [sec]})
started to rise again. We leave the experiments with other settings as a future exercise.

First, an unknown obstacle (i.e., depicted in Figure~\ref{Second environment/map used to test Pioneer 2DX} with O2)
was added to the test environment (room) and
the path planner application (online mode) was ran.
Second, another unknown obstacle (i.e., depicted  in Figure~\ref{Second environment/map used to test Pioneer 2DX} with O2)
was placed in the same test environment as before. As result the test environment contained
two unknown obstacles (i.e., O1 and O2).
Finally, for both of this scenarios the runtimes of the Pioneer 2DX
robot were measured by navigating from the initial location 
(depicted in Figure~\ref{Second environment/map used to test Pioneer 2DX} with letter R) to the final location
(depicted in Figure~\ref{Second environment/map used to test Pioneer 2DX} with letter D).
The results of these experiments are depicted in Table~\ref{Test results for the path planner application}.

Note that the obstacles depicted in the Pioneer simulator map
(Figure~\ref{Second environment/map used to test Pioneer 2DX}) are not
present in the path planner application---Figure~\ref{Path planer map available in the Java back-end}.
Thus the robot had to \emph{deal} with this obstacles in 
order to reach its target destination which was previously given 
(i.e., denoted with letter D in Figure~\ref{Path planer map available in the Java back-end}).

Figure~\ref{Path planer map available in the Java back-end} presents the
map of the room as it was \emph{modeled} in the steering 
application---blue map areas/yellow map areas 
represent non passable/passable map areas.
Note that this map did not contain the obstacles depicted in the map presented in
Figure~\ref{Second environment/map used to test Pioneer 2DX} at any time.
The 2D map is composed of squares which measure in reality 10 by 10 
centimeters. We found out experimentally that larger maps can be also used during our experiments.
We tested the the planning application by running the \emph{fast} A$^\ast$ algorithm in two 
different modes 
(depicted in Table~\ref{Test results for the path planner application} with M1 (first mode) and M2 (second mode)),  for more details see~\cite{Rosel:j}. 
Table~\ref{Test results for the path planner application} shows that first mode
is better suited with the map
depicted in Figure~\ref{Second environment/map used to test Pioneer 2DX} with one obstacle whereas the
second mode is better
suited for the map presented in Figure~\ref{Second environment/map used to test Pioneer 2DX} with both obstacles.
The first mode differs from second mode w.r.t. action limiters that are not added to the robot 
(not used during robot movement) in first mode.

\begin{table}[htb!]
\centering
\caption{Path planning with the \emph{fast} A$^\ast$ algorithm and two running modes}
\label{Test results for the path planner application}
\begin{tabular}{|c|l|c|c|} \hline
\textbf{Test run} &\textbf{Mode}   & \textbf{Map}                                                                  & \textbf{Time [sec]}  \\ \hline 
1                 &M1              &Figure~\ref{Second environment/map used to test Pioneer 2DX} with 1 obstacle   &47                    \\ \hline 
2                 &M2              &Figure~\ref{Second environment/map used to test Pioneer 2DX} with 1 obstacle   &45                    \\ \hline 
3                 &M1              &Figure~\ref{Second environment/map used to test Pioneer 2DX} with 2 obstacle   &61                    \\ \hline                  
4                 &M2              &Figure~\ref{Second environment/map used to test Pioneer 2DX} with 2 obstacle   &40                    \\ \hline

\end{tabular}
\end{table} 

Table~\ref{Test results for the path planner application} depicts the run-times for the 
two running modes on two real environments. 
In column four of Table~\ref{Test results for the path planner application} we observe 
that for the first environment (Figure~\ref{Second environment/map used to test Pioneer 2DX} with one obstacle) the best
run-time (47 seconds) is obtained with M1 selected and that for the second environment
(Figure~\ref{Second environment/map used to test Pioneer 2DX} with two obstacles)
the best run-time (40 seconds) is obtained 
with M2 turned on. 
Note that in Figure~\ref{Second environment/map used to test Pioneer 2DX}
we had two unknown obstacles (i.e., O1 and O2) which were added one after each other for each
of our experiments.
When running the \emph{fast} A$^\ast$ algorithm on the map depicted in 
Figure~\ref{Second environment/map used to test Pioneer 2DX} 
(containing only O1) with M1
the run-time increases w.r.t. M2 because the range of the
ultrasonic sensors was set tp 50 millimeters. 
Note that the sensors distance parameter for M1 was set to 50 millimeters whereas for M2
this value was set to 225 millimeters. 
Thus, the robot can make decisions earlier or later along the path.
As result the obstacle depicted in 
Figure~\ref{Second environment/map used to test Pioneer 2DX} (i.e., O1)
is detected later as compared to the detection of both obstacles
depicted in Figure~\ref{Second environment/map used to test Pioneer 2DX}
when the range of the sensors was increased to 225 millimeters.
Thus, the result is the addition of several recovery actions 
needed in order to recuperate the robot and
point him to the target destination.

However, when performing path planning with the map depicted
in Figure~\ref{Second environment/map used to test Pioneer 2DX} 
with two unknown obstacles using M2
the run-time decreases because the range of the
ultrasonic sensors was set to 225 millimeters and the result is that
the obstacles are detected earlier.
This removes additional recovery actions needed by the robot in order to
find a new obstacle avoiding path, thus time is not wasted.
We infer (with caution) from these results that the second mode is best suited for environments
with more unknown obstacles whereas the first mode is better suited for environments
with less unknown obstacles.

\section{Conclusion and Future Work} 
\label{Conclusion and Future Work} 
In this paper, we evaluated the A$^\ast$ algorithm and 
the \emph{fast} A$^\ast$ algorithm w.r.t. completness
and we shown that the \emph{fast} A$^\ast$ algorithm can 
be successfully used for indoor mobile robot navigation 
by using only data collected from ultrasonic sensors.
We built two software tools (for offline and online algorithm testing) which helped to tweak the 
used algorithms and to take further decisions based on this results.
The results obtained from comparing the A$^\ast$ algorithm and the \emph{fast}  A$^\ast$ algorithm (Section~\ref{The a vs a fast})
indicate a speed-up of two orders of magnitude w.r.t. the \emph{fast}  A$^\ast$ algorithm
inside our offline simulator.
The second mode used with the \emph{fast} A$^\ast$ algorithm is best suited for environments
with less unknown obstacles whereas the first mode is better suited for environments
with more than one unknown obstacles (in our experiments).
We are aware that further experiments are need in order to fully claim the above stated.
Additionally, we showed in our experiments that the \emph{fast} A$^\ast$ 
algorithm is complete (finds a path in due time). 
We leave the computation of its performance as a future exercise.

In future we want to further tweak the \emph{fast} A$^\ast$ algorithm and use other algorithms
with more complex unknown environments. 
We want to use more advanced path planning algorithms and
we want to combine input from multiple sensors (i.e., perform
fusion of data from several sources) which will give a more precise description of the environment.
Furthermore, we want to compute the performance of the used algorithms and compared them with each other.

\section*{Acknowledgements}
We want to express our gratitude to the anonymous reviewers for their constructive criticism.

\balance
\bibliographystyle{ieeetr}

\end{document}